\documentclass[runningheads]{llncs}

\usepackage{eccv}

\usepackage{bm}
\usepackage{epsfig}
\usepackage{graphicx}
\usepackage{amsmath}
\usepackage{amssymb}
\usepackage{microtype}
\usepackage{xcolor}
\usepackage{booktabs}
\usepackage{makecell}
\usepackage{caption}
\usepackage{tabularx}

\usepackage{tikz}
\usetikzlibrary{positioning, arrows.meta, fit, calc, decorations.pathreplacing}

\usepackage[accsupp]{axessibility}
\usepackage{hyperref}

\usepackage{orcidlink}

\makeatletter
\usepackage{xspace}
\DeclareRobustCommand\onedot{\futurelet\@let@token\@onedot}
\def\@onedot{\ifx\@let@token.\else.\null\fi\xspace}
\def\eg{\emph{e.g}\onedot}

\def\etc{\emph{etc}\onedot}
\makeatother

\begin{document}

\title{TinyHistory: Lightweight Video History Embeddings via Two-Stage Context Learning}

\titlerunning{TinyHistory}

\author{Lvmin Zhang\inst{1} \and
Shengqu Cai\inst{1} \and
Muyang Li\inst{2} \and
Chong Zeng\inst{1} \and
Beijia Lu\inst{3} \and
Anyi Rao\inst{4} \and
Song Han\inst{2} \and
Gordon Wetzstein\inst{1} \and
Maneesh Agrawala\inst{1}}

\authorrunning{L.~Zhang et al.}

\institute{$^1$Stanford University \quad
$^2$MIT \quad
$^3$Carnegie Mellon University \quad
$^4$HKUST}

\maketitle

\begin{abstract}
History context is central to autoregressive video generation, driving consistency and storytelling for both commercial models and personal use cases. For example, personal users, offline workflows, and individual-scale finetuning need to encode longer video histories under tight compute and memory budgets. We observe that content and identity consistency is an essential requirement, and that complete, uninterrupted history coverage together with content query and interpretation capabilities is broadly desired. We present TinyHistory, a lightweight history embedding learned through two-stage context learning. In the first stage, we pretrain the encoder on large-scale video data with a randomized frame query objective; in the second stage, we repurpose the pretrained encoder within an autoregressive video diffusion model to learn content-level consistency. As a result, we show that the learned lightweight embeddings achieve consistency comparable (by VLM, VBench, ELO, \etc) to heavier alternatives, while reducing training overhead and extending the encodable history length within a given memory budget. We conduct ablation studies to analyze the influence and trade-offs of each component.
\end{abstract}

\def\para#1{\vspace{0.25em}\noindent\textbf{#1.}}

\def\subpara#1{\noindent\emph{#1.}}

\section{Introduction}
\label{sec:intro}

Storytelling capability, narrative coherence, and context consistency are essential to the video generation community. The latest commercial models like Sora2~\cite{sora2}, Veo3.1~\cite{veo31}, and Seedance~2.0~\cite{seedance2} enable storyboard-based creation, scene planning, and dynamic camera work. Recent academic models have also focused on long video streaming~\cite{yin2024slow,huang2025selfforcing}, scene planning~\cite{guo2025lct,huang2024context,zhao2024moviedreamer,zheng2024videogen,zhou2024storydiffusion}, and video context consistency~\cite{huang2025conceptmaster,jiang2024videobooth,long2024videostudiogeneratingconsistentcontentmultiscene}. Among these directions, autoregressive models are a key paradigm for video storytelling, supporting native video continuation and storyboard streaming. Autoregressive models treat video history as context, and they face particular challenges when handling long-form content. These challenges affect both commercial-scale systems and personal use cases: for personal users, offline workflows, and individual-scale finetuning, encoding longer video histories is constrained by limited compute and memory.

A range of approaches have been explored to handle long video histories. A naive sliding window, which discards all distant-enough frames, maintains a fixed context length but loses long-range history. Compressed VAEs like LTXV~\cite{hacohen2024ltx} and DC-AE~\cite{chen2024deep} can produce more compact contexts, and hybrid approaches~\cite{bachmann2025flextok,jin2024pyramidal,zhou2025hitvideo} like FramePack~\cite{zhang2025framepack} operate at multiple levels, but at the cost of high-frequency image details. Another strategy is to reduce computation while preserving context length (\eg, using sparse~\cite{xi2025sparse, zhang2025spargeattn, zhang2025sta, xia2025trainingfree} or linear~\cite{cai2023efficientvit,xie2024sana,wang2020linformer,choromanski2020rethinking,yu2022metaformer,katharopoulos2020transformers} attention), though the context footprint in memory remains unchanged. Token merging~\cite{bolya2023tomesd,bolya2023tome} demonstrates that a higher merging rate leads to greater detail loss. These efforts indicate that the balance among history coverage, visual fidelity, and memory cost remains an open problem, and that the goal of a lightweight history representation is driven by the demands of different generative applications.

We observe that autoregressive video generation places three demands on a practical history embedding. First, content and identity consistency is an essential requirement: generative models need dense history features to maintain character identity, clothing, and scene layout across steps. Second, maximizing video context continuity needs the history to be represented in a complete and uninterrupted manner, covering the full temporal extent rather than relying on windowed or sparse keyframe sampling. Third, the embedding needs content query and interpretation capabilities, allowing different generation tasks to prioritize their most critical content within a feature manifold of constrained dimension.

We present TinyHistory, a lightweight history embedding learned through a two-stage context learning framework. The two stages are designed for the aforementioned demands: in the first stage, we pretrain the encoder on large-scale video data using a randomized frame query objective, to facilitate query capability and history coverage; in the second stage, we repurpose the pretrained encoder within an autoregressive video diffusion model on natural video data to establish content-level consistency. This approach allows the model to learn from the training data distribution how to prioritize different demands within the constrained feature manifold: \eg, models trained on game footage with repetitive scenes may prioritize scene appearance, while models trained on movies may tend to dedicate more features to face and character consistency.

Several challenges arise in learning such a lightweight history embedding. First, aligning a history representation with the generation model's hidden states requires careful design to avoid degradation and artifacts. We reuse the DiT's hidden manifold: the encoder outputs directly at the DiT's inner hidden states, bypassing the VAE bottleneck, to manipulate deep DiT features instead of the latent space. Second, training with full-length video contexts incurs a large computational cost. We observe that a useful indicator of the embedding's query and interpretation capabilities is its ability to attend to frames at arbitrary temporal positions, leading to a partial supervision strategy: the encoder is pretrained to query frames at randomly sampled time indices, ensuring dense coverage while avoiding processing the entire history at once. After pretraining on millions of videos, the encoder is integrated into the autoregressive system and repurposed for content-level consistency via low-rank adaptation.

Experiments show that the learned lightweight embeddings achieve content consistency, identity preservation, and user preference scores (measured by VLM evaluation, VBench, and ELO) comparable to heavier alternatives, while reducing training overhead and extending the encodable history length within a given memory budget. We compare with previous baselines and discuss the trade-offs among different approaches.

In summary, we (1) discuss the demands for a lightweight video history embedding in autoregressive generation, and the requirements arising from local and personal use cases; (2) present TinyHistory, a lightweight history embedding learned through a two-stage context learning framework, in which pretraining addresses query capability and history coverage, and joint repurposing within an autoregressive video diffusion model addresses content-level consistency; and (3) conduct extensive experiments across seven baseline paradigms, including quantitative evaluations, user studies, and ablation studies, to analyze the trade-offs of each component.

\section{Related Work}

\para{Autoregressive video diffusion and long videos}
Diffusion has become the driving force behind video synthesis, where a central challenge is length scaling.
Training-free length‑extension methods~\cite{qiu2023freenoise, lu2025freelong, ma2025freelongpp, zhao2025riflex, ruhe2024rolling} reschedule noise or re-balance temporal frequency to stretch pre-trained models beyond their training horizon.
A complementary thread blends diffusion with causal prediction: Diffusion Forcing~\cite{chen2025diffusion} and HistoryGuidance~\cite{song2025history} enable variable-horizon conditioning and stable long rollouts by noise injection.
These approaches are adapted in industrial systems such as SkyReels‑V2~\cite{chen2025skyreelsv2infinitelengthfilmgenerative} and Magi-1~\cite{teng2025magi1}.
Additionally, StreamingT2V~\cite{henschel2024streamingt2v} augments existing models with short- and long-term memory with randomized blending to generate longer videos.
FAR~\cite{gu2025longcontextautoregressivevideomodeling} studies long‑context AR with flexible RoPE~\cite{su2024roformer} decay and mixed short/long windows, while CausVid~\cite{yin2024slow} distills a bidirectional teacher to a few‑step causal generator.
StreamDiT~\cite{kodaira2025streamdit} combines multi-step distillation with a moving frame buffer and mixed partition training to generate results in real-time.
To mitigate error accumulation during AR generation, Self‑Forcing~\cite{huang2025selfforcing} simulates AR rollout during training, while its extensions~\cite{cui2025selfforcingpp, liu2025rollingforcing, yang2025longlive} further improve length generalization.

\para{Context learning and history representation}
Long-context persistence is crucial as we extend video generation beyond a few seconds.
One option is retrieval, which grounds generation in a persistent state.
WorldMem~\cite{xiao2025worldmem} and Context-as-Memory~\cite{yu2025contextasmemory} augment AR video world models with FoV-based history retrieval, while VMem~\cite{li2025vmem} indexes past views by surfels to retrieve the most relevant viewpoints.
Memory Forcing~\cite{huang2025memoryforcing} pairs geometry-indexed spatial memory with tailored training regimes to balance exploration and revisits.
Beyond fixed retrieval rules, Mixture-of-Contexts~\cite{cai2025moc} learns a dynamic sparse attention context router so that tokens attend only to the most salient chunks.
Similarly, MoGA~\cite{jia2025moga} and Holocine~\cite{meng2025holocine} also propose token-level sparse attending policies.
Pack-and-Force~\cite{wu2025packandforce} instead proposes a learnable context semantic retriever.
Image-level conditioning approaches such as IP-Adapter~\cite{ye2023ip} and its video extensions~\cite{ipadapterwan2025} inject CLIP vision embeddings as cross-attention conditions.
Orthogonal to retrieval, compression of history turns an unbounded context into a compact state.
FramePack~\cite{zhang2025framepack} compresses prior frames into a fixed‑size latent ``packed'' context.
Captain Cinema~\cite{xiao2025captaincinema} uses a similar compression for keyframes.
StateSpaceDiffuser~\cite{savov2025statespacediffuser} and Po et al.~\cite{po2025ssmworldmodel} swap quadratic attention for recurrent states to maintain long‑term memory.
TTTVideo~\cite{dalal2025oneminutevideogenerationtesttime} and LaCT~\cite{zhang2025lact} use light MLP test-time training layers as a learned context representation.

\para{Efficient video diffusion designs}
As we scale video generation to long horizons, large context windows are bottlenecked, driving a wave of efficient computational designs.
Kernel advances such as FlashAttention~\cite{dao2022flashattentionfastmemoryefficientexact, dao2023flashattention2} improve throughput.
Static or hardware-friendly sparse patterns include sliding/tiling 3D windows~\cite{zhang2025sta}, radial spatiotemporal masks~\cite{li2025radialattention}, and training-free head/pruning heuristics~\cite{xi2025sparse, yang2025svg2}.
Dynamic or learned pruning/routing further select salient token pairs or blocks~\cite{wu2025vmoba}, coarse-to-fine sparse token selection~\cite{zhang2025vsa}, Sage/SpargeAttention families~\cite{zhang2025sageattention, zhang2024sageattention2, zhang2025spargeattn}, blockified routing with cached search~\cite{xia2025trainingfree}, and progressive block carving~\cite{zhang2025jenga}.
Another line of work leverages compressing the latent space or token sequence: token merging and patch scaling~\cite{bolya2023tome, lee2024tokenmergevid}, compact/variable-rate tokenizers~\cite{bachmann2025flextok}, highly compressed latent space~\cite{hacohen2024ltx}, or multiscale pyramids with re-noising~\cite{jin2024pyramidal}.
SANA~\cite{xie2024sana} introduced a linear‑attention Diffusion Transformer for images, while SANA‑Video~\cite{chen2025sanavideo} extends this with block‑linear attention and a constant‑memory KV cache.

\section{Method}
\label{sec:method}

\begin{figure*}[t]
\centering
\includegraphics[width=\textwidth]{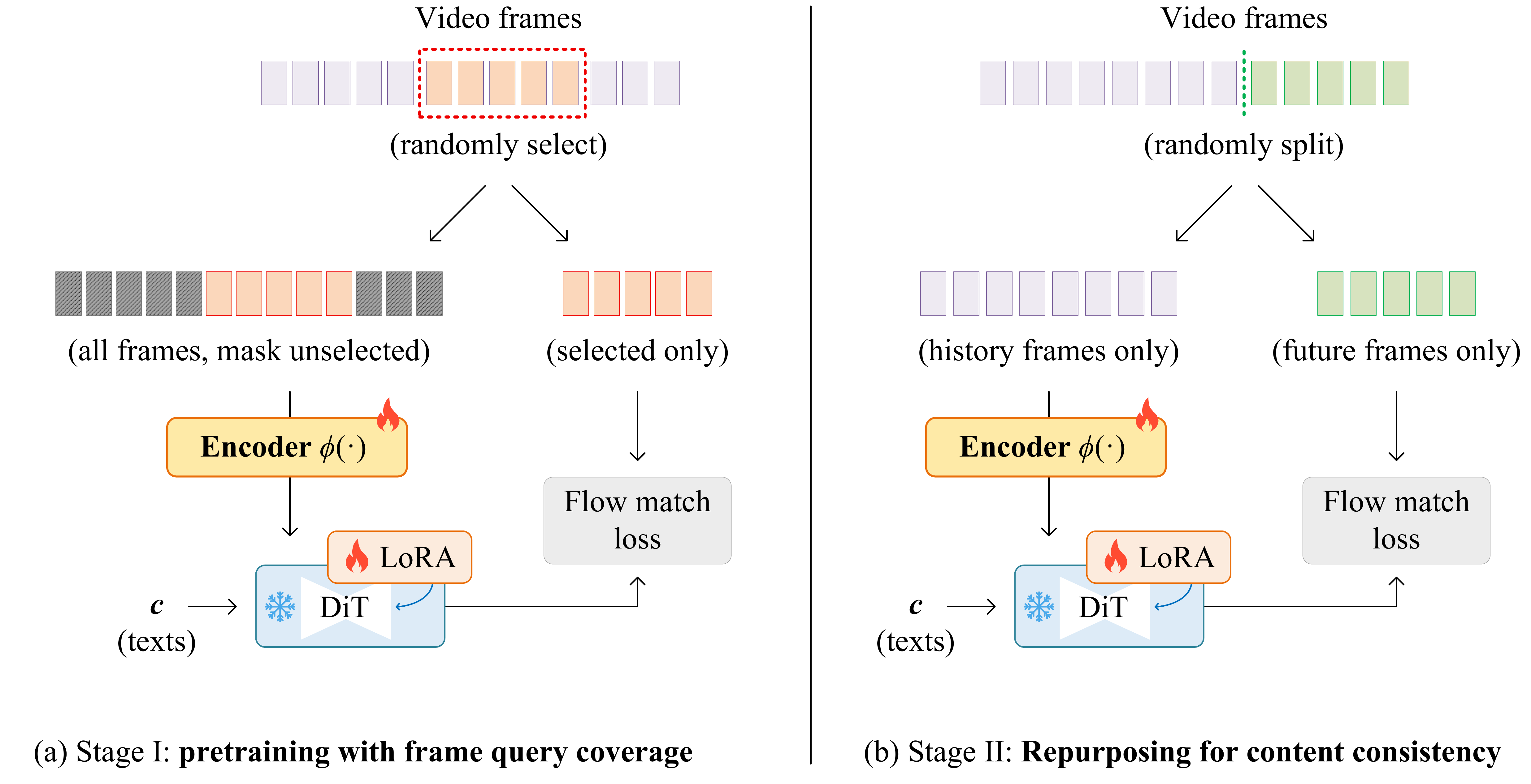}
\caption{Overview of TinyHistory. (a)~In Stage~I, the history encoder $\phi(\cdot)$ is pretrained on large-scale video data: random frame indices $\bm{\Omega}$ are selected as query targets, and the DiT learns to query these frames conditioned on the encoded history $\phi(\bm{H})$. (b)~In Stage~II, the pretrained encoder is integrated into an autoregressive video diffusion model and repurposed jointly (via LoRA) with natural video data to establish content-level consistency. Since the encoder is almost fully convolutional, new segments can be concatenated to the history embedding without recomputation.}
\label{fig:method}
\end{figure*}

We learn a history encoder $\phi(\cdot)$ through two-stage context learning (Fig.~\ref{fig:method}). The encoder maps a video history $\bm{H}$ into a lightweight embedding $\phi(\bm{H})$ that conditions an autoregressive DiT generator. Section~\ref{sec:encoder} details the encoder architecture. In Stage~I (Section~\ref{sec:stage1}), the encoder is pretrained on large-scale video data with a randomized frame query objective. In Stage~II (Section~\ref{sec:stage2}), the encoder is embedded into an autoregressive video diffusion model and repurposed to establish content-level consistency.

\para{Preliminaries} Unless otherwise noted, all ``frames'', ``pixels'', \etc, refer to latent concepts. We consider Diffusion Transformers (DiTs) such as Wan~\cite{wang2025wan} and HunyuanVideo~\cite{kong2024hunyuanvideo} with rectified-flow scheduling. The noisy latents $\bm{X}_{t_i}\in\mathbb{R}^{T \times H \times W \times C}$ are formed by
\begin{equation}
\bm{X}_{t_i} = (1 - t_i)\bm{X}_0 + t_i \bm{\epsilon}, \quad \bm{\epsilon} \sim \mathcal{N}\big{(}\mathbf{0}, \mathbf{I}\big{)},
\label{eq:flow}
\end{equation}
from clean latents $\bm{X}_0$, where $t_i\in(0,1]$ is the diffusion timestep. In autoregressive generation, the model conditions on a video history $\bm{H}\in\mathbb{R}^{T_h \times H \times W \times C}$. The generator $\bm{G}_\theta(\cdot)$ is trained by finding
\begin{equation}
\arg\min_\theta \mathbb{E}_{\bm{X}_0, \bm{H}, \bm{c},\bm{\epsilon}, t_i\sim \mathcal{L}(0,1)} \bigg\Vert (\bm{\epsilon} - \bm{X}_{0}) - \bm{G}_\theta\big{(}\bm{X}_{t_i}, t_i, \bm{c}, \bm{H}\big{)} \bigg\Vert_2^2,
\label{eq:diff}
\end{equation}
where $\bm{c}$ denotes conditions such as text prompts, and $t_i\sim \mathcal{L}(0,1)$ is the shifted logit-normal distribution~\cite{esser2024scaling}.

\subsection{History Encoder Architecture}
\label{sec:encoder}

\begin{figure}[t]
\centering
\includegraphics[width=0.8\linewidth]{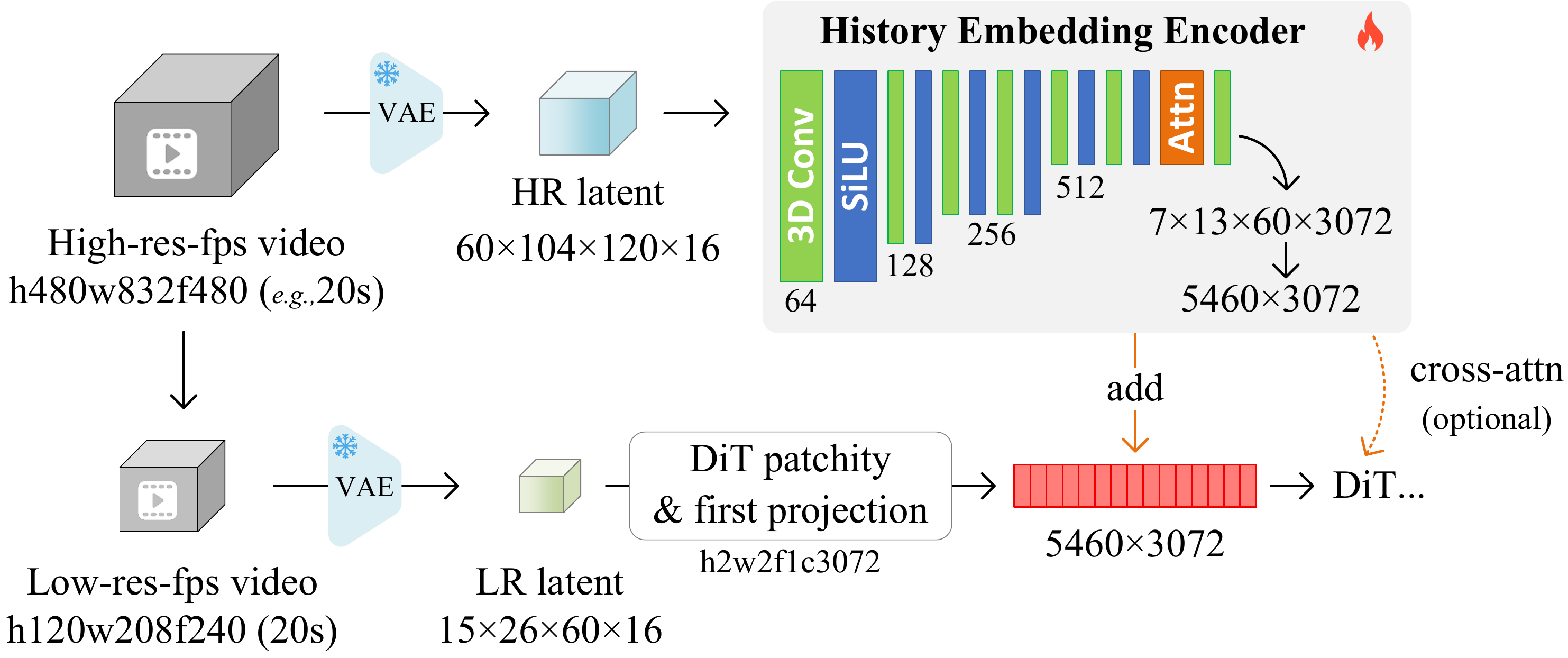}
\caption{History encoder architecture. The LR branch reuses the DiT's VAE, patchifier, and first-layer projection on a downsampled input. The HR branch encodes the original-resolution frames through a lightweight module. The two branches are summed after the DiT's first-layer projection, so the output operates in the DiT's inner hidden states rather than passing through the VAE channel bottleneck. Alternative architectures are discussed in ablation studies.}
\label{fig:arch}
\end{figure}

\looseness=-1
The history encoder $\phi(\cdot)$ maps a video history $\bm{H}$ of $T_h$ frames into a history embedding $\phi(\bm{H})$ that serves as context for the DiT generator. As shown in Fig.~\ref{fig:arch}, the encoder uses a dual-branch design. The first branch processes a low-resolution, low-frame-rate version of the history through the DiT's own VAE, patchifier, and first-layer projection, producing a coarse feature sequence. The second branch takes the original-resolution frames and passes them through a lightweight module (3D convolutions, SiLU activations, and attention layers) to produce a residual feature. The two branches merge via element-wise addition \emph{after} the DiT's first-layer projection. The encoder outputs are connected to the DiT's \emph{inner} hidden states (\eg, 3072 or 5120 channels) rather than being limited to the VAE channel bottleneck (\eg, 4, 16, or 64 channels), which is a key difference between this method and video compression methods like VAEs. Furthermore, an optional cross-attention refinement from the encoder's last hidden states to the DiT hidden states is discussed in ablation studies.

\subsection{Stage I: Pretraining with Frame Query Coverage}
\label{sec:stage1}

The first stage pretrains the history encoder $\phi(\cdot)$ on large-scale video data. The pretraining objective is a randomized frame query task that encourages the encoder to learn dense history features covering the full temporal extent.

Given a video history $\bm{H}$, we randomly sample a set of frame indices $\bm{\Omega}$. Frames at the selected indices are kept clean, while all remaining frames are masked with noise sampled from $\mathcal{L}(0.2, 1)$, resulting in a masked history $\texttt{mask}(\bm{H}, \bm{\Omega})$. The DiT learns to query the selected frames via a conditional mapping
\begin{equation}
\bm{G}_\theta : \{\phi(\texttt{mask}(\bm{H}, \bm{\Omega})), \bm{c}\} \mapsto \bm{H}_{[\bm{\Omega}]},
\label{eq:mapping}
\end{equation}
trained by finding
\begin{equation}
\arg\min_{\theta,\phi} \mathbb{E}_{\bm{H}, \bm{\Omega}, \bm{c},\bm{\epsilon}, t_i} \bigg\Vert (\bm{\epsilon} - \bm{H}_{[\bm{\Omega}]}) - \bm{G}_\theta\big{(}(\bm{H}_{[\bm{\Omega}]})_{t_i}, t_i, \bm{c}, \phi(\texttt{mask}(\bm{H}, \bm{\Omega}))\big{)} \bigg\Vert_2^2,
\label{eq:pretrain}
\end{equation}
where $\bm{G}_\theta(\cdot)$ is the DiT with low-rank adaptation (LoRA).

The randomness of $\bm{\Omega}$ is important: if only frames at fixed positions (\eg, the last few frames) are used as targets, the encoder can learn a cheating shortcut by dedicating all capacity to those positions while ignoring the rest of the history. Random selection forces the embedding to maintain dense coverage across the entire temporal extent. Since training on all frames at once incurs considerable overhead for minute-level histories even with efficient infrastructure, this partial supervision through $\bm{\Omega}$ provides a pretraining strategy that is both scalable and budget-friendly.

\subsection{Stage II: Repurposing for Content Consistency}
\label{sec:stage2}

The second stage integrates the pretrained encoder into an autoregressive video diffusion model and repurposes the system jointly to establish content-level consistency. Unlike Stage~I, the history $\bm{H}$ is now complete and noise-free, and no frame indices $\bm{\Omega}$ are involved. The DiT generator learns to produce the next video segment $\bm{X}$ conditioned on the history embedding
\begin{equation}
\bm{G}_\theta : \{\phi(\bm{H}), \bm{c}\} \mapsto \bm{X},
\label{eq:mapping2}
\end{equation}
by the diffusion objective
\begin{equation}
\arg\min_{\theta,\phi} \mathbb{E}_{\bm{X}_0, \bm{H}, \bm{c},\bm{\epsilon}, t_i} \bigg\Vert (\bm{\epsilon} - \bm{X}_{0}) - \bm{G}_\theta\big{(}\bm{X}_{t_i}, t_i, \bm{c}, \phi(\bm{H})\big{)} \bigg\Vert_2^2,
\label{eq:finetune}
\end{equation}
where newly initialized LoRA parameters are applied to adapt $\bm{G}_\theta(\cdot)$.

We note that the encoder $\phi(\cdot)$ is \emph{not} frozen during Stage~II: both the encoder and the LoRA-adapted DiT are jointly optimized. This distinguishes our approach from video compression and reconstruction research (\eg, VAEs, Codec, \etc), where the encoder is trained once and then locked. Because $\phi(\cdot)$ remains trainable, its features are repurposed from frame-level query targets (Stage~I) toward learning content-level consistency (Stage~II). The training data distribution then naturally shapes what the encoder prioritizes within the constrained feature manifold: for instance, models trained on game footage may dedicate more capacity to scene appearance, while models trained on cinematic data may prioritize face and character consistency.

\subsection{Autoregressive Inference}
\label{sec:inference}

At inference time, video is generated autoregressively: the model generates a segment $\bm{X}$, appends it to the history $\bm{H}$, and generates the next segment conditioned on the updated history embedding. Since the encoder is almost fully convolutional, the embedding for newly appended frames can be computed independently and concatenated to the existing embedding, without reprocessing the full history. This reduces the per-step encoding cost.

\section{Experiments}
\label{sec:exp}

\subsection{Experimental Details}
\label{sec:setup}

\para{Implementation}
We conduct Stage~I pretraining with 8$\times$H100 GPUs and Stage~II LoRA training with 1$\times$H100 or A100. 
The pipeline uses pure PyTorch with Accelerate, precomputed latents and conditions, and no gradient accumulation. 
The base models are Wan~2.1~14B~\cite{wang2025wan} and HunyuanVideo~12.8B~\cite{kong2024hunyuanvideo}, both flow-match DiTs at 480p resolution.
On a single 8$\times$A100-80G node, a smaller model (\eg, Wan~2.1~1.3B) at 480p achieves a batch size of approximately 64 for LoRA training with a window size of 8 (or batch size 32 with window size 16).
LoRA rank is 128 for all DiT adaptations. 
We use Wan~2.1~14B as the default; additional results with HunyuanVideo~12.8B are also reported.

\para{Hyper-parameters}
We denote encoding rates as $H{\times}W{\times}T$: for example, $4{\times}4{\times}2$ means reducing the latent height by 4$\times$, width by 4$\times$, and temporal length by 2$\times$.
The 3D convolution layers first reduce the temporal dimension, then the spatial dimensions via strided convolutions.
Hidden channels follow the pattern $64 \rightarrow 128 \rightarrow 256 \rightarrow 512$ and remain at 512 (\eg, in the attention layer) until a final $1{\times}1$ convolution projects the features to the DiT's inner hidden dimension (\eg, 3072 or 5120).

\para{Data preparation}
The training set consists of approximately 5 million internet videos, roughly half vertical short-form and half horizontal.
After quality filtering, the high-quality portion is captioned with Gemini-2.5-flash, and the remainder with QwenVL~\cite{wang2024qwen2}.
Captions are generated in storyboard format with timestamps; the prompt nearest to each timestamp is used for autoregressive training.

\para{Test set}
The test set comprises 1300 storyboard prompts generated by Gemini-2.5-pro and Gemini-3.1-pro, together with 4096 unseen videos independent from the training data.
All quantitative metrics are computed on videos generated from these prompts.

\subsection{Evaluation Metrics}
\label{sec:metrics}

\para{Gemini-3.1-pro VLM judgements}
Inspired by the VLM-as-judge idea in VBench~\cite{huang2023vbench}, we design four consistency dimensions with custom gate and sub-questions, and query Gemini-3.1-pro with uniformly sampled frames from each video.
For each dimension, a yes/no gating question first determines whether the aspect is present; if the gate passes, three yes/no sub-questions probe specific attributes.
The per-video score is the fraction of ``yes'' answers among the three sub-questions (four levels: 0, 0.33, 0.67, 1.0); the dataset score is the mean over all gated-in videos.
Full prompt templates are provided in the appendix.

\subpara{Character consistency (Char)}
The VLM is first asked ``is a clearly visible character present?''; if so, it is further asked whether the character's facial appearance, body structure, and overall identity remain coherent across frames.

\subpara{Scene consistency (Scene)}
The VLM is first asked ``is a visible background or environment present?''; if so, it is further asked whether the environment layout, lighting conditions, and background details remain stable.

\subpara{Object consistency (Obj)}
The VLM is first asked ``is a distinct non-person object prominently visible?''; if so, it is further asked whether the object's shape, color and texture, and physical plausibility remain consistent.

\subpara{Cloth consistency (Cloth)}
The VLM is first asked ``is a character with visible clothing present?''; if so, it is further asked whether the clothing color, style and design, and texture remain consistent.

\para{VBench metrics}
We additionally select three algorithmic metrics from VBench~\cite{huang2023vbench} that cover evaluation aspects distinct from the above VLM dimensions: text-video alignment, face identity, and motion dynamics.

\subpara{Semantic alignment (Semantic)}
The cosine similarity between ViCLIP~\cite{wang2023internvid} video and text embeddings, measuring how well the generated content aligns with the text prompt.

\subpara{Face identity (FaceID)}
Each frame's ArcFace~\cite{deng2019arcface} embedding is compared with the first frame's; the score is the fraction of frames whose cosine similarity exceeds a pre-defined value.

\subpara{Dynamic degree (Dynamic)}
RAFT~\cite{teed2020raft} optical flow is used to classify each video as dynamic or static; the score is the fraction of videos classified as dynamic.
This serves as a sanity check that history conditioning does not suppress motion.

\para{User preference (ELO)}
This is pairwise A/B preference tests and Bradley-Terry ELO ratings (base 1200, $K{=}32$).
Methods with obvious artifacts or severe inconsistency are excluded from the user study to reduce human workloads, marked ``/'' in the tables.

\subsection{Comparison Baselines}
\label{sec:baselines}

We organize baselines by seven architectural paradigms for handling video history, covering the spectrum from naive to learned approaches.
For each baseline, we implement the pipeline on the same Wan~2.1 base model and training data to ensure a controlled comparison.
Where a method has tunable hyperparameters, we align the context token length across methods for fairer comparison.
Detailed CTX/s derivations and implementation details for all methods are provided in the supplementary material.

\para{Sliding window}
The most recent 8 latent frames (approximately 2 seconds) are kept as clean latent context; all earlier history is discarded.
The DiT directly attends to these frames without any learned encoding (6240 tokens/s).

\para{Patchifier}
FramePack~\cite{zhang2025framepack} packs history frames into a fixed-size latent context by enlarging the patchifying projection kernel, so that temporally distant frames receive coarser spatial patches (2340 tokens\dag).

\para{Image planning + I2V}
QwenEdit~\cite{wu2025qwenimagetechnicalreport} takes the last $K$ generated keyframes as input and produces the next keyframe; all keyframes are generated first, then each is animated by Wan~2.1~I2V and concatenated.
We test $K{=}1$ and $K{=}2$; the history context is sparse (keyframe images only, no dense coverage).

\para{Retrieval-based memory}
Resembling WorldMem~\cite{xiao2025worldmem} and Context-as-Memory~\cite{yu2025contextasmemory}, this baseline estimates camera poses for all generated frames via COLMAP, predicts the next camera field-of-view (FOV) along the trajectory, and retrieves the history frames whose FOV is closest as context (1560 tokens/s; full history pool 4680 tokens\dag).

\para{Image embeddings}
Resembling IP-Adapter~\cite{ye2023ip} and video-IP-Adapter~\cite{ipadapterwan2025}, this baseline uniformly samples 2 frames per second from the history, extracts a CLIP Vision embedding for each, and injects the concatenated embeddings into the DiT via a projection layer ($\sim$512 tokens/s).

\para{Feature compression}
Resembling APT~\cite{choudhury2025apt} and ToMe~\cite{bolya2023tome}, this baseline computes each history latent token's cosine similarity with its six $T{\times}H{\times}W$ neighbors (up, down, left, right, front, back); tokens exceeding a threshold $\alpha$ are removed and their features projected into the remaining neighbors ($\sim$512 tokens/s).
The operation is mathematically equivalent to a large-kernel convolution.

\para{External VAE}
The LTX-Video VAE~\cite{hacohen2024ltx} (see also DCAE~\cite{chen2024deep}) encodes the video history at a higher compression rate than Wan~2.1's native VAE; a new randomly initialized linear projection then learns to map the compressed latent to the DiT's hidden dimension.
An alternative is to adopt LTX-Video directly as the base model, but a fair comparison would then require all methods to use the same LTX-Video backbone; this is discussed in the supplementary material.

\subsection{Results}
\label{sec:results}

\begin{table*}[t]
\centering
\caption{Comparison with baselines on content consistency. CTX/s denotes the history context length per second on Wan~2.1 at 480p. Best results in \textbf{bold}, second best \underline{underlined}. ``\dag'': CTX covers the entire history rather than per second. ``/'': Excluded from user study due to severe artifacts.}
\label{tab:baseline}
\resizebox{\textwidth}{!}{
\begin{tabular}{lcc cccc ccc c}
\toprule
& & & \multicolumn{4}{c}{Gemini-3.1-pro VLM $\uparrow$} & \multicolumn{3}{c}{VBench $\uparrow$} & User $\uparrow$ \\
\cmidrule(lr){4-7} \cmidrule(lr){8-10} \cmidrule(lr){11-11}
Method & Paradigm & CTX/s & Char & Scene & Obj & Cloth & Semantic & FaceID & Dynamic & ELO \\
\midrule
Sliding Window & Naive & 6240 & 46.65 & 57.02 & 48.52 & 45.00 & 22.55 & 57.87 & \textbf{90.61} & 1090 \\
FramePack~\cite{zhang2025framepack} & Patchifier & 2340\dag & 57.92 & 76.04 & 67.09 & 63.83 & 23.65 & 62.01 & \underline{88.16} & 1182 \\
QwenEdit~\cite{wu2025qwenimagetechnicalreport}+I2V ($K{=}1$) & Image planning & Sparse & 58.60 & 65.87 & 62.80 & 58.09 & 23.61 & 62.21 & 79.14 & / \\
QwenEdit~\cite{wu2025qwenimagetechnicalreport}+I2V ($K{=}2$) & Image planning & Sparse & 74.67 & 81.81 & 72.61 & 75.54 & 25.46 & 68.81 & 79.49 & 1205 \\
FOV Retrieval (resem.~\cite{xiao2025worldmem,yu2025contextasmemory}) & Retrieval-based Memory & 1560 (4680\dag) & 55.80 & \textbf{88.17} & 59.38 & 55.83 & 21.68 & 65.78 & 83.29 & / \\
CLIP Vision (resem.~\cite{ye2023ip,ipadapterwan2025}) & Image Embeddings & 512 & 49.06 & 60.37 & 53.59 & \textbf{93.73} & 22.17 & 59.89 & 83.57 & / \\
Token Merging (resem.~\cite{bolya2023tome,choudhury2025apt}) & Feature Compression & $\sim$512 & 51.09 & 63.73 & 55.30 & 48.09 & 23.14 & 60.61 & 83.96 & / \\
LTX-VAE~\cite{hacohen2024ltx}+Proj & External VAE & 780 & 62.70 & 69.72 & 64.69 & 66.82 & 22.98 & 67.01 & 82.17 & 1147 \\
\midrule
TinyHistory ($4{\times}4{\times}2$) & Two-stage Learning & 195 & \underline{80.19} & 72.21 & \underline{74.38} & 77.40 & \underline{25.74} & \underline{70.71} & 84.31 & \underline{1262} \\
TinyHistory ($2{\times}2{\times}2$) & Two-stage Learning & 780 & \textbf{85.90} & \underline{87.11} & \textbf{85.45} & \underline{87.74} & \textbf{26.84} & \textbf{73.05} & 86.41 & \textbf{1332} \\
\bottomrule
\end{tabular}
}
\end{table*}

\para{Comparison with baselines}
Three baselines each top exactly one dimension in Table~\ref{tab:baseline} --- FOV Retrieval in Scene (88.17), CLIP Vision in Cloth (93.73), Sliding Window in Dynamic (90.61) --- but all three suffer non-trivial degradation on multiple dimensions.
TinyHistory ($2{\times}2{\times}2$) leads Char, Obj, Semantic, FaceID, and ELO, and ranks second in Scene and Cloth, at 780 tokens/s.
The Sliding Window retains full-length latent (6240 tokens/s) yet reports the lowest Char (46.65) and FaceID (57.87), suggesting that longer temporal coverage is important for these metrics.
At the same 780 tokens/s, TinyHistory ($2{\times}2{\times}2$) reports higher scores than LTX-VAE+Proj across all dimensions (Char 85.90 vs 62.70): learning projections for adapting external VAE latent space can cause local minima and quality degradation, and TinyHistory's encoder bypasses the latent channel bottleneck of the external VAE by directly connecting to the DiT inner hidden states.
TinyHistory ($4{\times}4{\times}2$) at 195 tokens/s still leads all external baselines in five dimensions, at the cost of Scene (72.21, rank 5) and Cloth (77.40, rank 3) with $4{\times}$ spatial rate; the gap with $2{\times}2{\times}2$ (Scene 72.21 vs 87.11) illustrates the encoding rate trade-off examined in Section~\ref{sec:ablation}.

\para{Equivalent per-second context length}
CTX/s in Table~\ref{tab:baseline} denotes the number of history context tokens per second on Wan~2.1 at 480p (derivations in the supplementary material).
The full-length baseline is 6240 tokens/s.
Methods whose context does not grow with video length are marked ``\dag'': FramePack packs all history into a fixed 2340-token context, and FOV Retrieval retrieves 1560 tokens/s with the full history clamped to 4680 tokens.
Image planning methods (QwenEdit) operate on sparse keyframe images and are marked ``Sparse'' as their cost is not directly comparable in token units.
TinyHistory ($4{\times}4{\times}2$) operates at 195 tokens/s, a $32{\times}$ reduction from the full-length baseline, while TinyHistory ($2{\times}2{\times}2$) at 780 tokens/s matches the per-second cost of LTX-VAE+Proj.

\begin{figure*}[t]
\includegraphics[width=\textwidth]{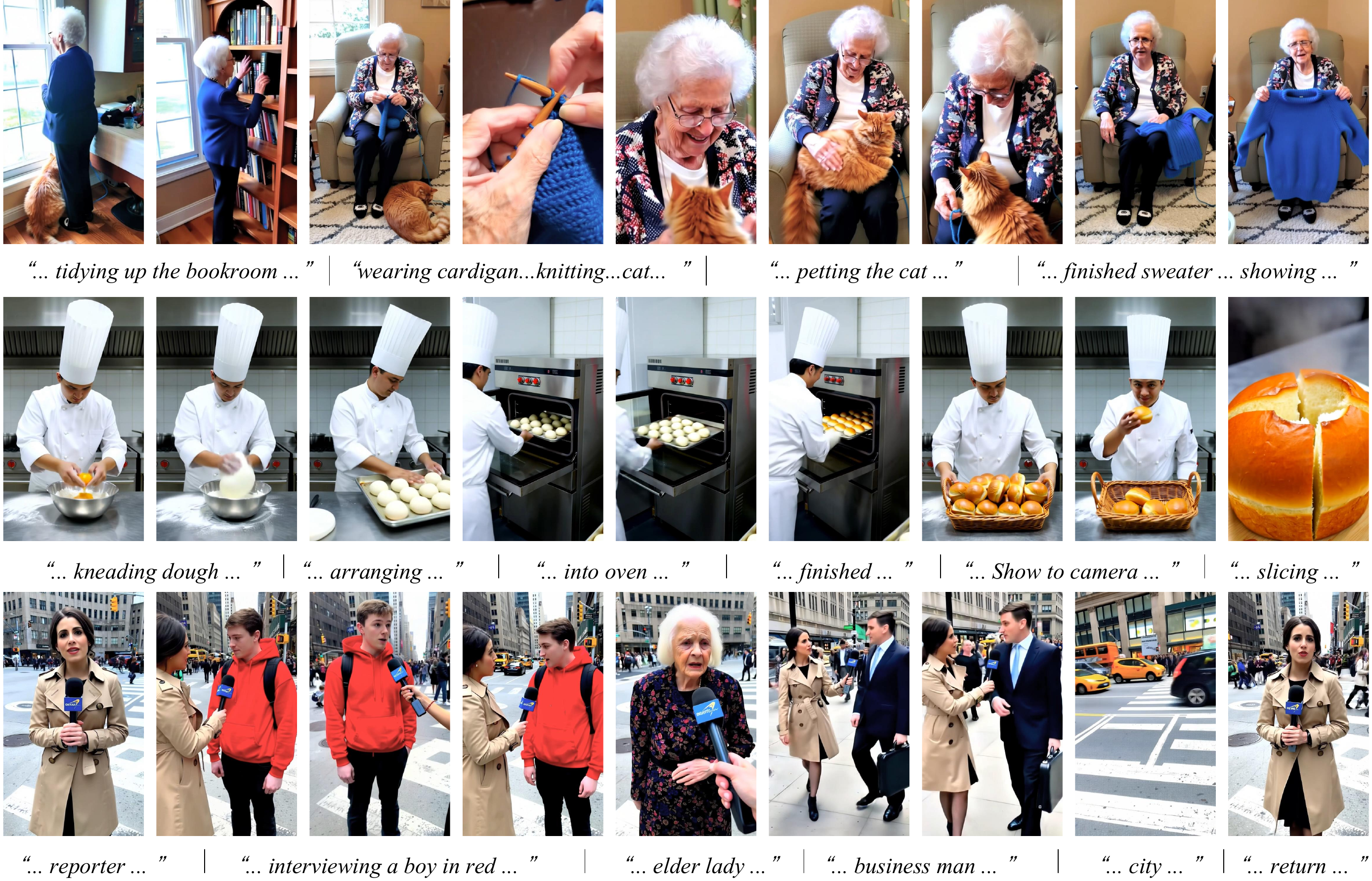}
\caption{Qualitative results on storyboard generation. Each row shows frames sampled from an autoregressively generated video driven by a storyboard of text prompts. This batch of results uses HunyuanVideo as base model. The history encoder maintains character identity, clothing, and scene layout across shots with diverse prompts. Storyboards are written by external language models.}
\label{fig:qual}
\end{figure*}

\para{Qualitative results}
Figure~\ref{fig:qual} shows storyboard-driven generation results, where each row is an autoregressively generated video driven by a sequence of text prompts.
Across diverse prompts that introduce new activities, camera angles, and environments, this method preserves character identity, clothing, and scene layout across shots.

\subsection{Ablation Studies}
\label{sec:ablation}

Table~\ref{tab:ablation} presents ablation results addressing four design questions: (1)~is Stage~I pretraining necessary? (2)~must the encoder be trainable in Stage~II? (3)~are both branches necessary? (4)~how does encoding rate affect consistency?

\begin{figure}[t]
\includegraphics[width=\linewidth]{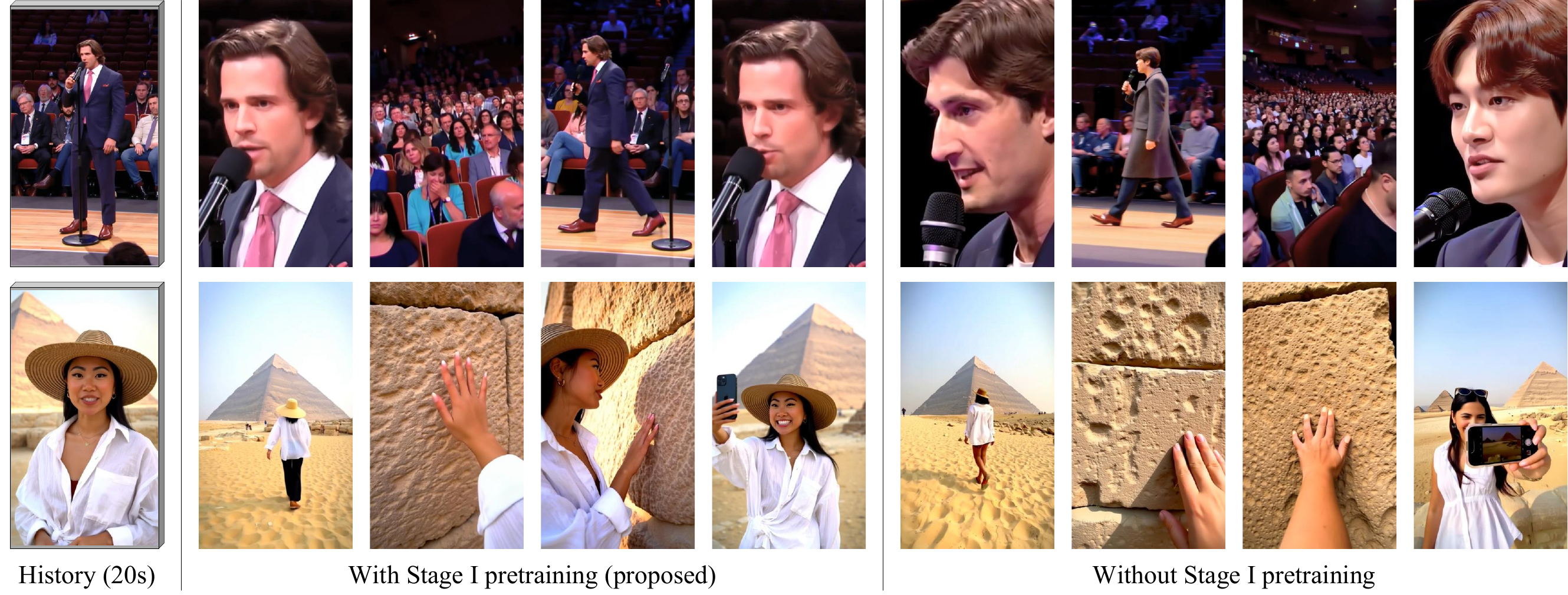}
\caption{Influence of Stage~I pretraining on generation consistency. Given the same video history, the model with Stage~I pretraining (left) preserves facial features, clothing, scene layout, and camera coordination, while the model without pretraining (right) fails to attend to relevant history content, resulting in identity and object inconsistency. Both models are trained for 100k steps.}
\label{fig:pretrain}
\end{figure}

\begin{table}[t]
\centering
\caption{Ablation study. All variants use Wan~2.1 as the base model and are trained for 100k steps. CTX/s denotes the history context length per second on Wan~2.1 at 480p. Best results in \textbf{bold}, second best \underline{underlined}.}
\label{tab:ablation}
\resizebox{0.85\linewidth}{!}{
\begin{tabular}{lc cccc ccc c}
\toprule
& & \multicolumn{4}{c}{Gemini-3.1-pro VLM $\uparrow$} & \multicolumn{3}{c}{VBench $\uparrow$} & User $\uparrow$ \\
\cmidrule(lr){3-6} \cmidrule(lr){7-9} \cmidrule(lr){10-10}
Variant & CTX/s & Char & Scene & Obj & Cloth & Semantic & FaceID & Dynamic & ELO \\
\midrule
Without Stage~I ($4{\times}4{\times}2$) & 195 & 52.64 & 52.84 & 52.53 & 50.37 & 20.50 & 58.38 & \textbf{95.69} & / \\
Frozen Encoder ($4{\times}4{\times}2$) & 195 & 61.42 & 62.69 & 58.33 & 58.39 & 22.31 & 61.82 & \underline{92.00} & / \\
Only LR ($4{\times}4{\times}2$) & 195 & 74.29 & 65.06 & 67.03 & 76.12 & 23.55 & 67.28 & 89.17 & 1223 \\
Without LR ($4{\times}4{\times}2$) & 195 & 67.73 & 58.01 & 64.57 & 54.86 & 22.70 & 65.36 & 90.72 & / \\
\midrule
Proposed ($4{\times}4{\times}2$) & 195 & 80.19 & 72.21 & 74.38 & 77.40 & 25.74 & 70.71 & 84.31 & 1262 \\
Proposed ($2{\times}2{\times}2$) & 780 & 85.90 & \underline{87.11} & \textbf{85.45} & \textbf{87.74} & \textbf{26.84} & \textbf{73.05} & 86.41 & \underline{1332} \\
Proposed ($2{\times}2{\times}4$) & 390 & \underline{90.24} & 81.02 & 77.06 & 79.35 & 25.14 & 71.59 & 85.04 & 1281 \\
Proposed ($2{\times}2{\times}1$) & 1560 & \textbf{93.78} & \textbf{92.59} & \underline{84.56} & \underline{85.54} & \underline{26.01} & \underline{72.20} & 88.90 & \textbf{1342} \\
\bottomrule
\end{tabular}
}
\end{table}

\para{Stage~I: Influence of pretraining}
Without Stage~I pretraining, only Stage~II with random initialization is insufficient to learn effective history representations due to local minima and difficulty in converging.
Char drops from 80.19 to 52.64 and FaceID from 70.71 to 58.38.
Dynamic rises to 95.69, the highest in Table~\ref{tab:ablation}; weaker history conditioning imposes less constraint on motion, a consistency-dynamics trade-off visible across all degraded variants.
Figure~\ref{fig:pretrain} shows that the model without pretraining exhibits identity and consistency degradation, while the pretrained model maintains consistent characters and scenes.

\para{Stage~II: Frozen vs.\ trainable encoder}
Freezing the encoder after Stage~I --- treating it as a locked encoder --- drops Char from 80.19 to 61.42, indicating that Stage~I features must be repurposed through joint Stage~II training.
\textit{Frozen Encoder} ranks between \textit{Without Stage~I} and \textit{Only LR} (Char 52.64 < 61.42 < 74.29): pretrained initialization contributes but is insufficient without adaptation.
Dynamic remains high at 92.00, consistent with the pattern that weaker consistency signals yield freer motion.

\para{Branch ablation}
Removing the LR branch (\textit{Without LR}, Char 67.73) degrades more than removing the HR branch (\textit{Only LR}, Char 74.29), suggesting that DiT manifold alignment through the LR path may contribute more to the overall feature structure.
\textit{Without LR} also exhibits Dynamic 90.72: degraded history embeddings impose less constraint on motion, yielding randomly dynamic transitions with poor consistency.
The full dual-branch design (Char 80.19) improves over both, indicating the two paths provide complementary features.

\para{Encoding rate}
The rate sweep is largely monotonic, but with diminishing returns: $2{\times}2{\times}2$ at 780 tokens/s achieves comparable quality to $2{\times}2{\times}1$ at 1560 tokens/s (ELO 1332 vs 1342) at half the context cost.
$4{\times}4{\times}2$ vs $2{\times}2{\times}4$ (195 vs 390 tokens/s) isolates spatial from temporal rate: $2{\times}2{\times}4$ reports Char 90.24 vs 80.19, indicating that $4{\times}$ spatial rate discards more identity-relevant detail than $4{\times}$ temporal rate.

\subsection{Extensions and Compatibility}
\label{sec:extensions}

The history embedding design is modular and compatible with several orthogonal enhancements. Table~\ref{tab:extensions} reports quantitative results when each extension is added on top of the base TinyHistory ($4{\times}4{\times}2$) configuration.

\begin{table}[t]
\centering
\caption{Extensions and compatibility. Each row adds one extension to the base TinyHistory ($4{\times}4{\times}2$). All extensions are orthogonal and can be combined.}
\label{tab:extensions}
\resizebox{0.6\linewidth}{!}{
\begin{tabular}{l ccc cc}
\toprule
& \multicolumn{3}{c}{Gemini-3.1-pro VLM $\uparrow$} & \multicolumn{2}{c}{VBench $\uparrow$} \\
\cmidrule(lr){2-4} \cmidrule(lr){5-6}
Configuration & Char & Scene & Obj & Semantic & FaceID \\
\midrule
Proposed ($4{\times}4{\times}2$) & 80.19 & 72.21 & 74.38 & 25.74 & 70.71 \\
\;+ Sliding Window & 85.75 & \textbf{82.51} & 78.41 & 25.75 & 71.94 \\
\;+ Cross-Attention & 87.01 & 75.32 & 80.75 & 25.54 & 72.72 \\
\;+ Multiple Encoders & \textbf{88.25} & 79.50 & \textbf{83.33} & \textbf{25.91} & \textbf{73.37} \\
\;+ KV-Cache & 75.80 & 65.28 & 70.51 & 24.89 & 63.61 \\
\bottomrule
\end{tabular}
}
\end{table}

\begin{figure}[t]
\centering
\begin{minipage}[t]{0.48\linewidth}
  \centering
  \includegraphics[width=\linewidth]{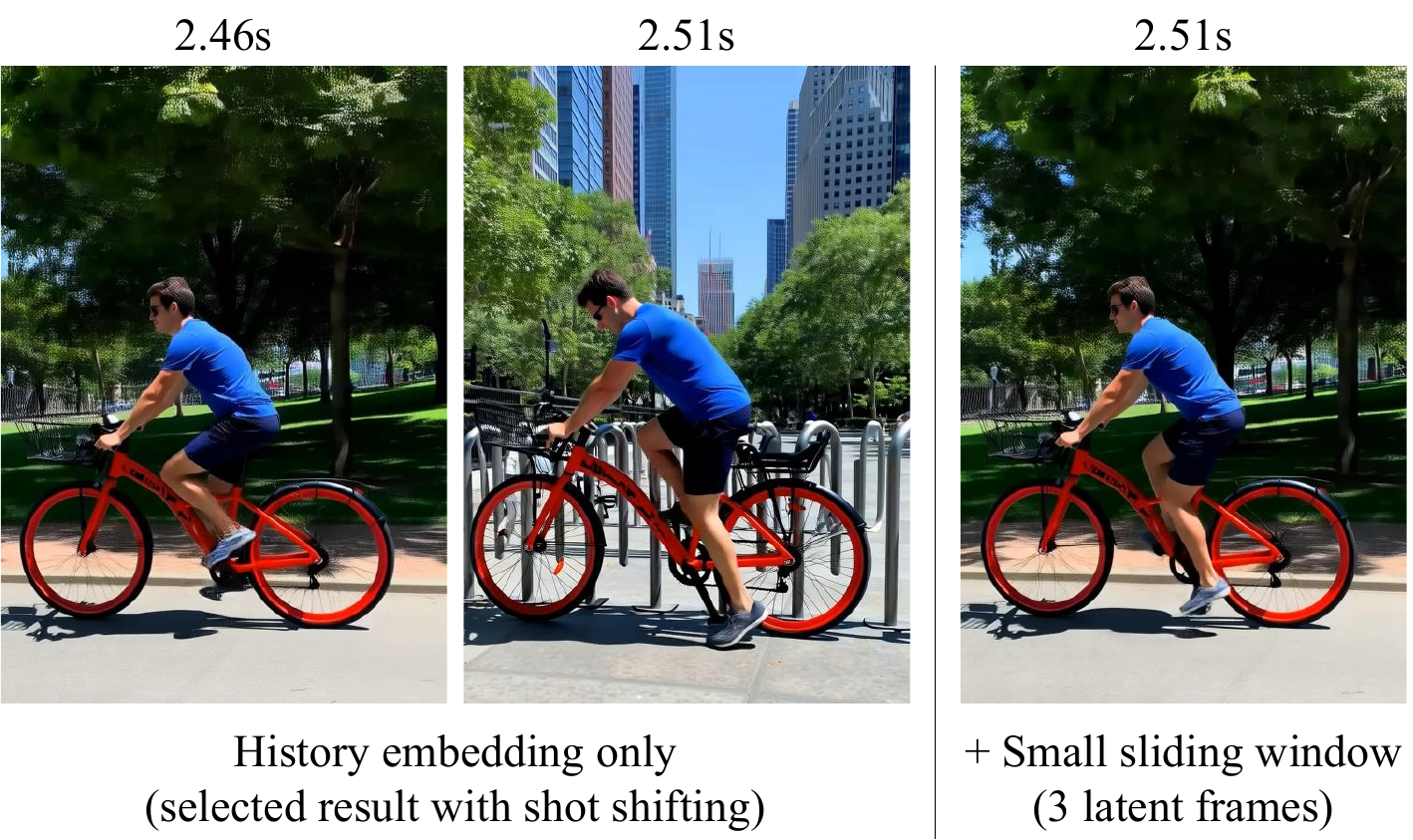}
  \caption{Adding a small sliding window (3 latent frames). Without the sliding window, consecutive generation steps may exhibit camera shot shifts. The sliding window provides a short overlap that encourages temporal continuity.}
  \label{fig:sli}
\end{minipage}
\hfill
\begin{minipage}[t]{0.48\linewidth}
  \centering
  \includegraphics[width=\linewidth]{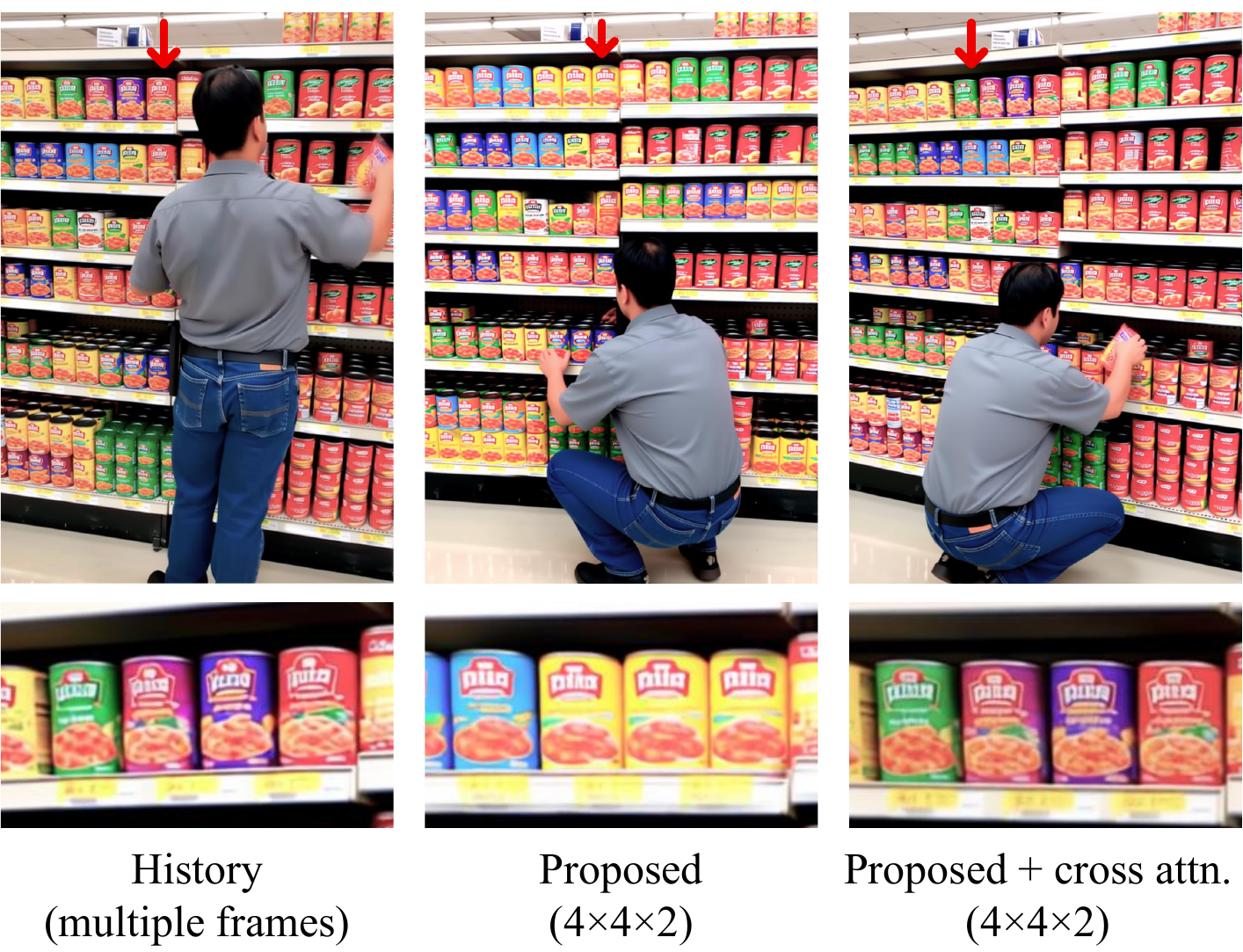}
  \caption{Adding cross-attention from the history encoder's last hidden states to the DiT. In detail-oriented scenarios (\eg, the items on shelves), cross-attention provides additional consistency at the cost of increased computation.}
  \label{fig:cross}
\end{minipage}
\end{figure}

\para{Sliding window}
Adding a 3-frame latent overlap ($\sim$0.75 seconds) between consecutive generation steps reduces the chance of camera shot shifts when the DiT receives only the history embedding as context.
Scene consistency increases by 10.3 points (72.21$\to$82.51), the largest single-extension gain in Table~\ref{tab:extensions}; Char and Obj also increase by 5.6 and 4.0 points respectively (Fig.~\ref{fig:sli}).

\para{Cross-attention enhancement}
Adding an extra cross-attention connection from the encoder's last hidden states to the DiT layers allows performing spatial queries over history features and improves fine-grained consistency.
Obj increases by 6.4 points (74.38$\to$80.75) and Char by 6.8 points, at the cost of additional computation per DiT block (Fig.~\ref{fig:cross}).

\begin{figure}[t]
\centering
\includegraphics[width=0.65\linewidth]{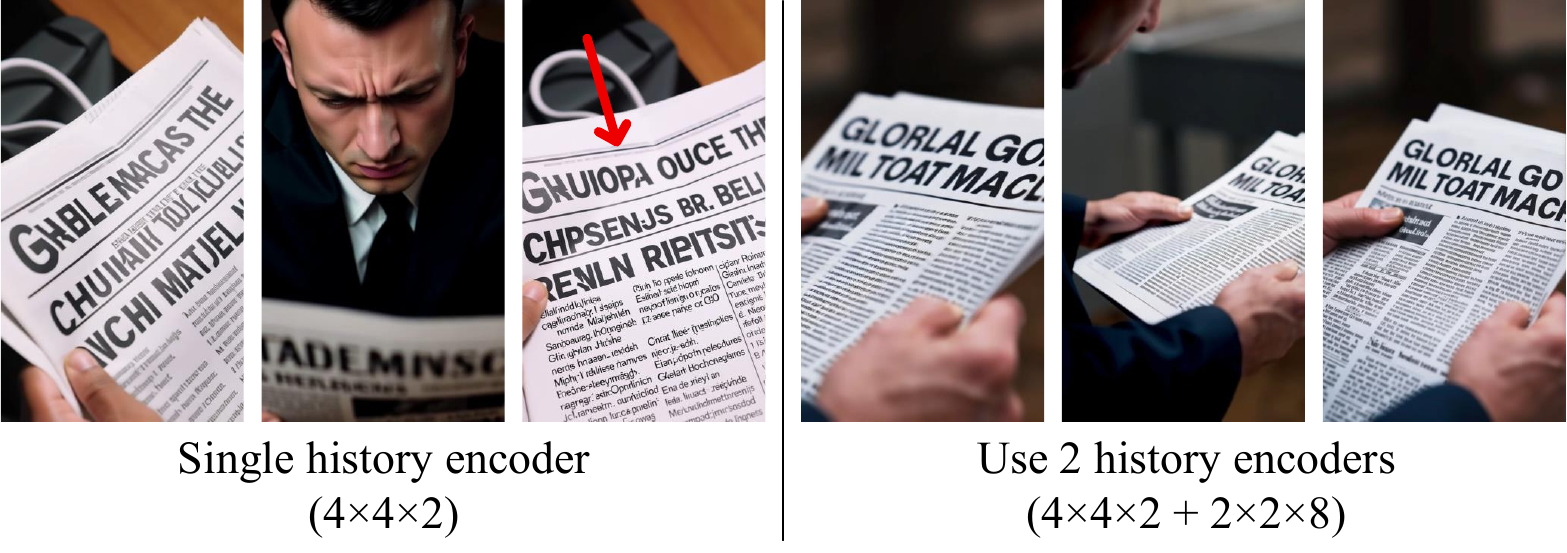}
\caption{Using two history encoders with different encoding patterns ($4{\times}4{\times}2$ and $2{\times}2{\times}8$). The second encoder prioritizes spatial detail over temporal coverage, enabling fine-grained features such as text on newspapers to be maintained across generation steps, at the cost of a longer context.}
\label{fig:mixing}
\end{figure}

\para{Multiple encoders}
A second encoder with a complementary encoding pattern ($2{\times}2{\times}8$, prioritizing spatial detail over temporal coverage) provides measurable gains: Char $+$8.1, Obj $+$9.0, FaceID $+$2.7 (Fig.~\ref{fig:mixing}).
The cost is a longer context ($195 + 195 = 390$ tokens/s total), as the two encoders' embeddings are concatenated before appending to the DiT context.

\para{KV-Cache compatibility}
Because the history embedding is fixed once encoded, its key-value pairs can be cached across diffusion steps, avoiding redundant attention computation.
This trades a quality decrease (FaceID $-$7.1, Scene $-$6.9) for inference speedup, and is orthogonal to causal inference acceleration methods such as CausVid~\cite{yin2024slow} and Self-Forcing~\cite{huang2025selfforcing}.

\section{Conclusion}
\label{sec:conclusion}

This paper presents TinyHistory, a two-stage context learning framework for lightweight video history embeddings in autoregressive generation.
In Stage~I, the encoder is pretrained with a randomized frame query objective on large-scale video data to establish dense history coverage; in Stage~II, the encoder is jointly repurposed with the DiT generator on natural video data to learn content-level consistency.
Experiments on Wan~2.1 and HunyuanVideo show that this method leads five of eight evaluation dimensions (Char, Obj, Semantic, FaceID, ELO) and ranks second in two others (Scene and Cloth), using shorter context than heavier alternatives.
Ablation studies confirm that pretraining, joint encoder training, and dual-branch design each contribute to the final quality, and that the encoding rate trade-off follows a diminishing-returns pattern.
The architecture is modular: sliding window overlap, cross-attention, multiple encoders, and KV-caching can each be added independently.

\bibliographystyle{splncs04}
\bibliography{db_video,db_efficiency}
\end{document}